%% file: main.tex
\documentclass[runningheads]{llncs}

\usepackage{eccv}

\usepackage{eccvabbrv}

\usepackage{graphicx}
\usepackage{booktabs}

\usepackage{multirow}
\usepackage{derivative}
\usepackage{algorithm}
\usepackage{algpseudocode}
\usepackage{wrapfig}
\usepackage{subcaption}
\usepackage[accsupp]{axessibility}  

\usepackage{hyperref}

\usepackage{orcidlink}

\begin{document}

\title{Weight-Space Mixture-of-Experts for Implicit Neural Representation Classification} 

\titlerunning{Weight-space mixture-of-experts}

\author{Stanislaw Janik\inst{1}, Michal Byra\inst{1,2}}

\authorrunning{S. Janik and M. Byra}

\institute{Institute of Fundamental Technological Research, \\ Polish Academy of Sciences, Warsaw, Poland \and
Samsung AI Center, Warsaw, Poland\\
\email{\{sjanik,mbyra\}@ippt.pan.pl}}

\maketitle

\begin{abstract}

Implicit Neural Representations (INRs) encode signals as the weights of a coordinate-based neural network and have recently been proposed as an alternative domain for downstream learning. While promising, classification directly in weight space remains challenging due to the high dimensionality and complex structure of INR parameters. Furthermore, the way discriminative information is distributed across INR weights remains poorly understood.
We propose a hierarchical Mixture-of-Experts (HMoE) Transformer that processes INR weights using conditional computation aligned with the structure of the underlying implicit network. Coupled with a meta-learning framework that shapes INR parameters for downstream tasks, our model achieves \textit{state-of-the-art} accuracy across standard benchmarks, ranging from low-resolution datasets to high-resolution ImageNet-1K.
To gain insight into how INRs encode discriminative information, we develop weight-space attribution and pruning methods that identify parameters most relevant for classification. These analyses reveal how class-specific structure emerges within INR layers and support the suitability of MoE architectures for weight-space learning.
Our approach advances both the performance and interpretability of weight-space classifiers.

\keywords{image classification \and implicit representations  \and explainability  \and mixture-of-experts \and weight-space learning}

\end{abstract}

\input{sec/1_intro}

\input{sec/2_related_work}

\input{sec/3_methods}

\input{sec/4_experiments}

\input{sec/5_conclusion}

\bibliographystyle{splncs04}
\bibliography{main}

\input{supp}

\end{document}

%% file: sec/1_intro.tex
\section{Introduction}
\label{sec:intro}

Implicit neural representations (INRs) encode signals using the weights of multilayer perceptrons (MLPs), offering a powerful alternative to traditional grid-based representations. Initially developed for image synthesis and compression, INRs have recently been explored for learning directly in weight-space \cite{essakine2024we}. In this paradigm, an image is represented by the parameters of an implicit network, and downstream models operate on these weights—rather than on pixel arrays—to perform tasks such as classification. This viewpoint provides a unified framework for pattern recognition, since diverse signals (e.g., images, audio) can all be mapped to INR parameters and processed in a consistent manner. Moreover, INRs are highly flexible and can be trained to jointly represent multiple quantities, opening promising avenues for multimodal processing and analysis. Additionally, because INRs encode data in network parameters rather than explicit signal samples, they offer potential advantages for privacy-preserving deep learning, for example in federated learning settings or biomedical applications where sharing raw data is restricted.

However, processing INR weights is challenging for several reasons. First, internal symmetries in MLPs complicate the design of effective weight-space methods \cite{godfrey2022symmetries}. For example, neurons within a layer can be permuted without affecting the represented function \cite{ainsworthgit,grigsby2023hidden}, which requires classifiers to respect these symmetries, ideally through equivariance or invariance. Second, the large number of parameters in typical MLPs demands models capable of handling high-dimensional weight vectors and, in many cases, compressing them. Third, training INR-based systems requires fitting a separate INR per data sample, which becomes computationally expensive for large-scale datasets \cite{ma2024implicit}.

The end-to-end Meta Weight Transformer (MWT) was proposed to address several of the challenges outlined above \cite{gielisse2025end}. In this framework, meta-learning is used to obtain a consistent initialization for the target dataset and to impose a more organized structure on the INR weights. After adaptation, the INR neurons are converted into tokens and processed by a Transformer classifier. The classifier and the meta-network are trained jointly, allowing the classification loss to backpropagate through the Transformer and shape the weight distribution of the meta-INR, promoting a discriminative alignment of the INR weights.

In this work, we take a step further and introduce a new classification architecture for INR weight-space learning. We propose a hierarchical Mixture-of-Experts (HMoE) model that processes INR weights more effectively than a plain Transformer and integrates naturally with the meta-learning framework used to adapt the INR parameters. Mixture-of-Experts (MoE) architectures consist of multiple specialized feed-forward experts together with a routing mechanism that directs each token to only a subset of them. This conditional computation is well suited for INR weight-space, where different groups of parameters encode distinct functional roles across layers and frequency components. Our motivation is consistent with broader findings in neural networks: neurons exhibit specialization \cite{olah2020zoom,hanni2024mathematical}, activation-based methods reveal structured response patterns \cite{simonyan2013deep,nguyen2016synthesizing}, class-relevant information in INRs can concentrate in specific components \cite{dupont2022data,kim2023generalizable}, and pruning studies show that coordinated subnetworks (“circuits”) carry most task-relevant information \cite{han2015learning,frankle2020linear,frankle2018lottery}. Our HMoE architecture leverages these insights through a two-stage design aligned with INR structure.

The main contributions of this work are as follows:

\begin{itemize}

\item We propose a novel approach to weight-space learning that combines meta-learning with a HMoE Transformer. Our results show that the MoE architectures are particularly well suited for operating in INR weight space. 

\item We achieve \textit{state-of-the-art} (SOTA) results across multiple image-classification benchmarks, ranging from low-resolution datasets such as MNIST \cite{mnist}, Fashion-MNIST \cite{fashionmnist}, and CIFAR-10 \cite{cifar10} to higher-resolution datasets including Imagenette \cite{imagenette} and ImageNet-1K \cite{imagenet}, demonstrating the effectiveness of our approach.

\item We develop weight-space explainability methods, including gradient based attribution and structured pruning, to analyze how classification-relevant information is distributed within the implicit network.

\item We conduct an ablation study to better understand the design choices of our approach.

\end{itemize}

%% file: sec/2_related_work.tex
\section{Related work}

\subsection{Classification based on implicit networks}

Classifying INRs requires operating directly on MLP weights, which exhibit strong permutation and scale symmetries that challenge conventional classifiers.

\noindent \textbf{Equivariant methods.}
Several architectures enforce invariance or equivariance to neuron permutations.
Representative examples include DWS-Net~\cite{dws_net}, permutation-equivariant neural functionals~\cite{perm_eq_neural_functionals}, and ScaleGMN~\cite{scale_equivariant_graph_metanetworks}. 
Graph-based approaches~\cite{lim2023graph,kofinasgraph} and permutation-equivariant Transformers~\cite{zhou2023neural} treat INRs as graphs, but typically scale poorly to wider networks. 
Probe-based methods such as ProbeGen~\cite{kahanadeep} avoid these symmetries by probing the INR as a black box, though they ignore useful structure in the weight space.

\noindent \textbf{Structured weight-space learning.}
Meta-learning provides an alternative by shaping INR parameters for downstream tasks. 
Functa and Spatial Functa~\cite{dupont2022data,bauer2023spatial} learn latent embeddings that modulate a shared implicit network for reconstruction. Instance Pattern Composers~\cite{kim2023generalizable} modulate INR layers via hypernetworks, while INR2VEC~\cite{ramirez2024deep} learns compact embeddings that reconstruct the original function.  
Most closely related to our work, the Meta Weight Transformer (MWT)~\cite{gielisse2025end} employs a MAML and MetaSGD based framework~\cite{maml} that jointly adapts the INR parameters and trains a Transformer classifier. By allowing the classification loss to influence the INR adaptation process, MWT encourages the learned weight landscape to organize in a way that improves downstream classification performance.

\subsection{Explainable image classification}

Explainability in vision models has been extensively studied.
Perturbation based methods such as RISE~\cite{Petsiuk2018rise} and meaningful perturbations~\cite{fong2017interpretable}, activation-based approaches such as Grad-CAM~\cite{selvaraju2017grad}, and gradient-propagation techniques including SmoothGrad~\cite{smilkov2017smoothgrad} and LRP~\cite{bach2015pixel} form three broad families of attribution methods. 
However, attribution for models that operate directly in INR weight space remains unexplored. 
We extend gradient-based attribution to this domain and combine it with structured weight pruning.

\subsection{Mixture-of-Experts}

Mixture-of-Experts architectures combine multiple feed-forward experts with sparse routing~\cite{shazeer2017outrageously}, enabling specialization while maintaining computational efficiency.
MoEs have been successfully applied to large vision and language models~\cite{riquelme2021scaling,shen2023scaling}, where experts naturally learn to handle distinct semantic or structural patterns.
However, their potential in INR weight space remains unexplored, despite the fact that INR parameters exhibit strong layer-wise structure, frequency decomposition, and naturally separable functional roles.
These properties make weight-space INRs an ideal setting for conditional computation.
We introduce a hierarchical MoE architecture that explicitly leverages this structure, enabling the model to specialize across both layers and tokens.

%% file: sec/3_methods.tex
\section{Methods}

\begin{figure*}[t]
      \centering
    \includegraphics[width=0.9\linewidth]{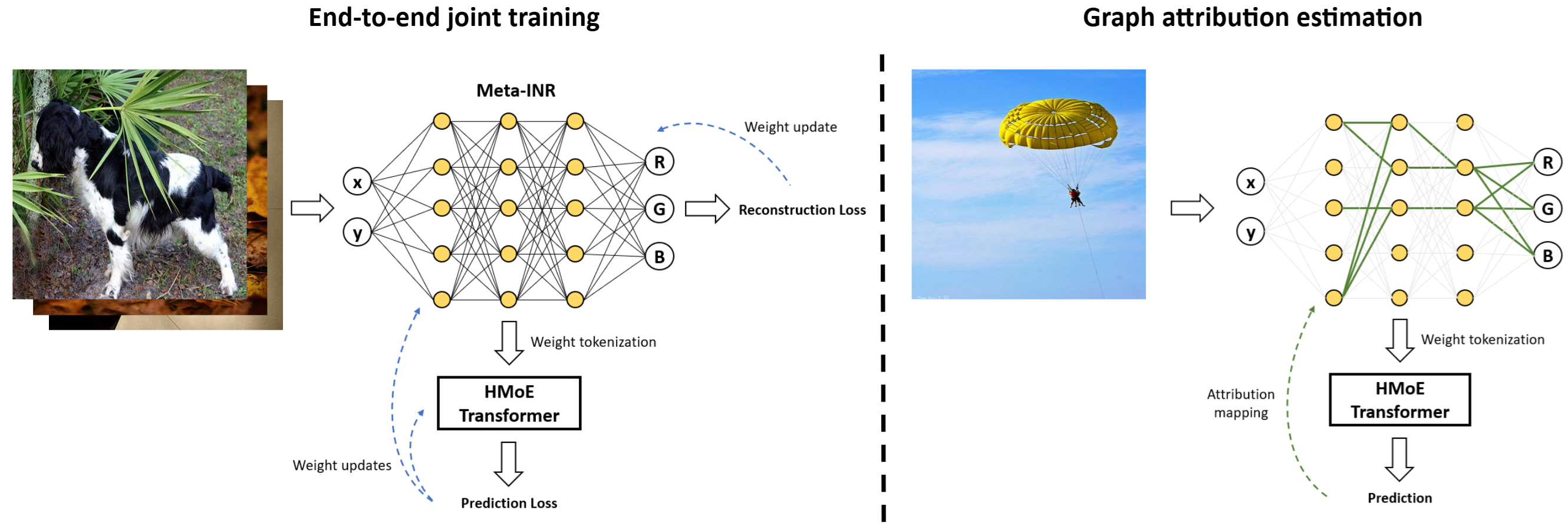}
    \caption{Overview of the proposed framework. End-to-end training jointly optimizes the meta-INR and the hierarchical Mixture-of-Experts (HMoE) classifier through reconstruction and classification losses (left). After training, our attribution pipeline computes weight-space importance scores, identifies relevant computational subcircuits, and explains classification decisions via graph-based attribution (right).}
    \label{fig:approach}
\end{figure*}

Our framework is summarized in Fig.~\ref{fig:approach}. We introduce the HMoE Transformer classifier designed to process the weights of implicit neural representations. The full system is trained end-to-end in a meta-learning setup, where the weight-space classifier and the meta-network that initializes the INR are optimized jointly. To analyze the resulting models, we additionally develop a gradient based attribution method that identifies class specific circuits within the INR parameters, providing insight into which parts of the weight space contribute to each classification decision. The subsections below describe each component of the framework.

\subsection{Classification in weight-space}

Conventional image classification models operate directly on pixel arrays. In contrast, weight-space classification represents each image as an INR and feeds the INR parameters to a downstream classifier. In our setting, an image is represented by a MLP $f_{\phi}$ that maps pixel coordinates to RGB intensities:
\[
f_{\phi}(p) = I \in \mathbb{R}^3, \qquad p \in [-1,1]^2,
\]
where $\phi$ denotes all INR weights and biases.  The classifier $c(\,\cdot\,)$ receives $\phi$ as input and predicts the corresponding image class. Classification is therefore performed entirely in weight-space, and the classifier must utilize the structural properties of INR parameters.

\subsection{End-to-end classification framework}

Our approach is based on the end-to-end framework that jointly trains a meta-learned INR and a weight-space Transformer classifier \cite{gielisse2025end}. Instead of fitting each INR from scratch, we determine a shared initialization $\theta$ and a per-parameter learning-rate vector $\alpha$, following MAML/Meta-SGD formulation \cite{maml}. For each training image, the INR is adapted for $k$ inner-loop steps using the reconstruction loss $\mathcal{L}_{\mathrm{rec}}$, producing an instance-specific parameter set $\phi$.

Importantly, the classification loss $\mathcal{L}_{\mathrm{cls}}$ is backpropagated to the INR, allowing the classifier to influence how the INR evolves during adaptation. This coupling encourages the meta-INR to learn weight structures that are informative for classification.

For classification, only the hidden layers of the INR are processed. Each hidden layer is a linear map with weight matrix $W \in \mathbb{R}^{n \times n}$ and bias $b \in \mathbb{R}^{n}$. The bias is folded into the weight matrix using a padded input, yielding $W' \in \mathbb{R}^{(n+1)\times n}$. Each output neuron corresponds to a token containing its $n+1$ incoming weights, giving a token dimension $D=n+1$. An INR with width $n$ and $L$ hidden layers thus produces $M = nL$ tokens, each in $\mathbb{R}^{D}$. Because INR tokens have no inherent positional order, a learned positional encoding is added to the tokens. The classifier operates on the difference $\phi - \theta$, which emphasizes the task-specific changes arising during the adaptation. A Transformer processes the resulting sequence, and the final token representations are average-pooled and passed through a linear classifier to produce class logits.

\subsection{Weight-space Mixture-of-Experts}

A standard Transformer treats all INR-weight tokens as exchangeable vectors. 
Even with positional encodings, the model applies the same feed-forward 
transformation to every token and therefore cannot exploit the structured 
organization of INR weights: neurons belong to specific layers, layers encode 
different frequency components, and discriminative information is often carried 
by coordinated groups of parameters rather than isolated weights. This motivates 
replacing the shared feed-forward block with a MoE module that 
introduces conditional computation and expert specialization.

\begin{figure}[!b] 
    \centering

    \includegraphics[width=7cm]{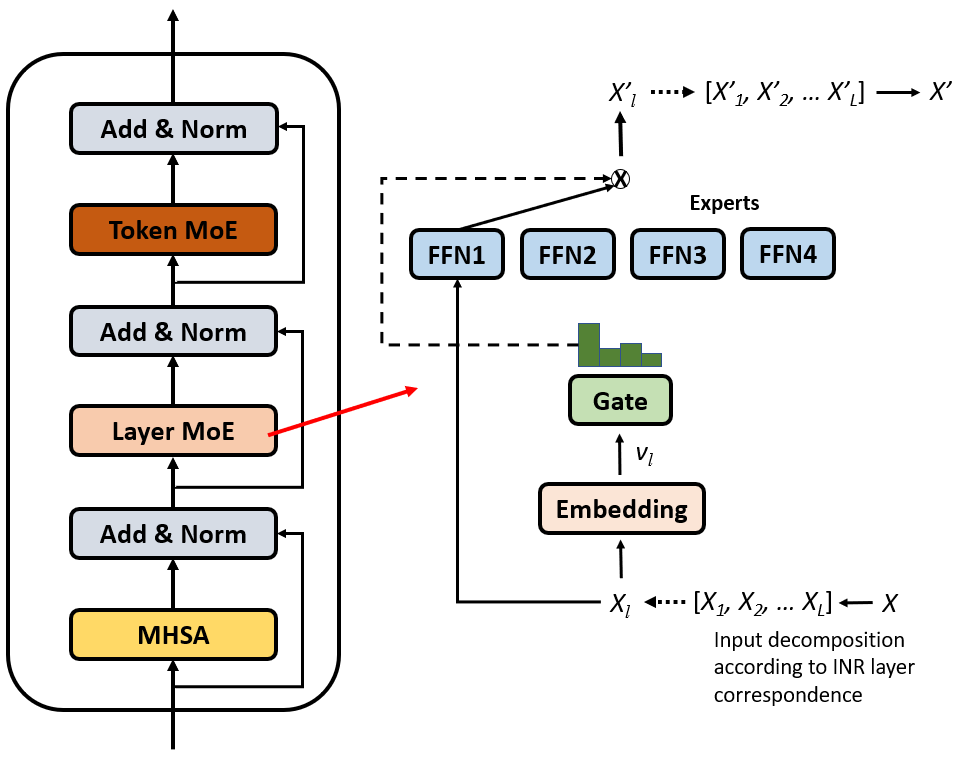}

\caption{The proposed hierarchical Mixture-of-Experts (HMoE) Transformer block.
The INR weights are tokenized according to their originating INR layer and processed by two MoE stages: a Layer MoE, which routes tokens based on layer-level structure, and a Token MoE, which performs fine-grained expert routing across all tokens.}
    \label{fig:moe}

\end{figure}

Our HMoE Transformer substitutes each feed-forward block with a two-stage sparse 
Mixture-of-Experts module, see Fig.~\ref{fig:moe}:

\paragraph{Stage 1: Layer-wise MoE.}

Let $X \in \mathbb{R}^{M \times D}$ be the sequence of INR-weight tokens. We regroup tokens by INR layer,
\[
X \rightarrow (X_1, \dots, X_L), \qquad X_{\ell} \in \mathbb{R}^{n \times D}.
\]
For each layer $\ell$, we compute a routing vector:
\[
g_{\ell} = \operatorname{softmax}(W_g v_{\ell}), 
\qquad
v_{\ell} = \frac{1}{n}\sum_{i=1}^n X_{\ell}[i],
\]
where $v_\ell$ summarizes all tokens in the layer. The top-$k$ experts are selected via $\mathrm{TopK}(g_{\ell}, k)$, and all tokens in that layer are processed by the selected experts:
\[
X_{\ell}' = \sum_{j \in \mathrm{TopK}(g_{\ell}, k)} g_{\ell,j} \, E_j(X_{\ell}).
\]
After all layer-specific token sets are processed by their selected experts, the 
outputs are concatenated in layer order: 
\[
X' = \mathrm{concat}(X_1', \dots, X_L'),
\]
forming the output of the layer-wise MoE stage.  

\paragraph{Stage 2: Token-wise MoE.}

In the second stage, the model applies token-wise MoE with conditioning from Stage 1. For a token $x \in X_\ell$ from layer $\ell$, the routing score is computed as: 
$$s= \text{softmax}(W_sx + g_\ell B),$$ where $g_\ell$ is the routing vector produced in Stage 1 and $B \in \mathbb{R}^{E_L \times E_T}$ is a learned projection mapping layer-level routing signals into the token-level expert space, with $E_L$ and $E_T$ denoting the number of experts at each stage. Each token is then processed by its top-$k$ experts.

\[
x' = \sum_{j \in \mathrm{TopK}(s, k)} s_{j} \, E_j(x).
\]

Our hierarchical design captures coarse layer-dependent structure in stage~1 and finer token-level specialization in stage~2, yielding a richer and more expressive alternative to a shared feed-forward layer. Our design is motivated by the fact that INR layers encode distinct frequency components of the signal, forming coherent functional blocks~\cite{siren,fourier_features,yuce2022structured}.  Routing entire layer-level token groups to a sparse top-$k$ subset of experts preserves this structure while enabling specialization across layers.

\subsection{Weight-space attribution}

Interpretability of weight-space classifiers remains largely unexplored. In contrast to standard vision models, where maps are typically defined over image pixels, our classifier operates solely on INR parameters; therefore, pixel-space attribution methods cannot be applied directly. To address this gap, we introduce Gradient-Weighted Class Activation Graph (Grad-CAG),  a weight-space analogue of Grad-CAM \cite{selvaraju2017grad}.

Given a classifier output $c(\phi)$ and a target class $i$, we compute:
\begin{equation}
    S^{(i)}_j \;=\; 
    \left| \phi_j \, \frac{\partial c_i(\phi)}{\partial \phi_j} \right|,
\end{equation}
assigning an importance score to each INR weight. This score measures the first-order sensitivity of the class logit to perturbations in $\phi_j$.

Grad-CAG provides a class-specific ordering of INR weights, enabling several downstream analyses. First, weights with large importance scores are presumed to belong to INR circuits that encode information useful for the target classification. These scores support a principled pruning strategy, allowing us to investigate how removing important or unimportant weights affects INR-based image reconstruction quality. Although Grad-CAG is a local, first-order measure, our quantitative and qualitative experiments show that it captures functionally meaningful structure in INR weight space and offers the first interpretability tool for weight-space classification.

%% file: sec/4_experiments.tex
\section{Experiments}
\label{sec:exp}
\noindent \textbf{Implementation.} For fair comparison with prior work, we adopt training settings aligned with \cite{gielisse2025end}. Our default classifier is a 10-block Transformer in which every standard Transformer block is replaced with the proposed hierarchical MoE block. Unlike \cite{gielisse2025end}, we follow the standard Transformer design and set the FFN dimension to four times the token dimension. The meta-network is a SIREN with four hidden layers. Each expert is implemented as a two-layer FFN, and gating modules are implemented as single linear layers. 

We evaluate several variants of our method. \textbf{HMoE-WT} and \textbf{HMoE-MWT} denote models trained without and with  injection of the classification loss into the meta-learning part. Both use a SIREN with hidden dimension 128. \textbf{HMoE-MWT-L} is a larger model using a SIREN with hidden dimension 256. 

We train all models using AdamW \cite{loshchilov2017decoupled} with the learning rates reported in \cite{gielisse2025end} and a batch size of 16, except for the HMoE-MWT-L variant, which uses a batch size of 4 due to memory constraints.

In addition to the mean-squared reconstruction loss and cross-entropy classification loss, we include a balancing loss to ensure stable expert utilization, averaging this term across all MoE modules in the Transformer \cite{cai2025survey}. Unless otherwise stated, the reconstruction, classification, and balancing losses are weighted by 1, 0.01, and 0.1, respectively.

We find that MoE-based models benefit from longer training and therefore train all models for 40 epochs unless specified otherwise. All variants employ sparse TopK routing with $k{=}1$ and four experts in both the layer-wise and token-wise stages.Since only one expert is activated per token, the computational cost remains comparable to a dense model, while the additional experts increase the total parameter capacity. Experiments were conducted on a workstation equipped with a NVIDIA RTX 4090 GPU. Additional training details are provided in the supplementary materials. Our code is available at https://github.com/stasiek-j/HMoE-MWT.

\subsection{Classification of INRs}

We conduct a thorough evaluation of our method using established classification benchmarks. First, we evaluate the  performance on low resolution datasets. Second, we examine the performance on challenging large scale datasets.

 \subsubsection{Low-resolution datasets}

Following prior work, we evaluate our approach on MNIST \cite{mnist}, Fashion-MNIST \cite{fashionmnist}, and CIFAR-10 \cite{cifar10}.  The meta-learning inner-loop employs 6 gradient updates. SIREN INR, with $\omega$ of 10, is trained using full default coordinate grid.

Table \ref{tab:performance_toydatasets} compares our approach with prior weight-space classifiers across the three benchmark datasets. Our method achieves good performance, surpassing the original MWT baselines \cite{gielisse2025end} and all competing INR and graph based models such as DWS \cite{dws_net}, NFN \cite{perm_eq_neural_functionals} and ScaleGMN \cite{scale_equivariant_graph_metanetworks}. Notably, HMoE-MWT achieves the highest accuracy of 99.06\% on MNIST, 90.72\% on Fashion-MNIST and 65.01\% on CIFAR-10. 

Importantly, even without meta-learning, our HMoE-WT variant substantially outperforms the standard WT, narrowing the gap to MWT. Moreover, HMoE-MWT surpasses the MWT-L baseline that uses higher-capacity SIRENs even when trained for the same extended schedule (vs 10 epochs originally). Interestingly, unlike \cite{gielisse2025end}, we do not observe accuracy gains from increasing the SIREN width, further emphasizing the importance of the classifier architecture. Overall, our results demonstrate the architectural advantages of our hierarchical MoE design and highlight that conditional expert specialization is particularly well suited for weight-space learning.

The authors of \cite{gielisse2025end} also reported a large-scale MWT model with 20 transformer blocks trained for 40 epochs with data augmentations, achieving 64.7\% accuracy on CIFAR-10. In comparison, our more compact 10-block HMoE-MWT already surpasses this result, demonstrating that architectural design has a stronger impact than merely increasing model dimensionality. Furthermore, scaling our model to 20 transformer blocks and applying augmentations further improved performance, achieving a new SOTA accuracy of 69.11\% on CIFAR-10.

\begin{table}[t]
\centering
\caption{Comparison of weight-space classifiers on MNIST, Fashion-MNIST and CIFAR-10 datasets. Our method outperforms the previously proposed methods, highlighting the usefulness of Mixture-of-Expert architectures for learning in weight spaces. The asterisk * indicates our version of the MWT-L model trained for 40 epochs. }
\footnotesize

\begin{tabular}{p{3cm} ccc}
\toprule

& \multicolumn{3}{c}{Classification Accuracy (\%)}\\
\textbf{Method} & \textbf{MNIST} & \textbf{Fashion-MNIST} & \textbf{CIFAR-10} \\
\midrule
MLP \cite{scale_equivariant_graph_metanetworks}  & 17.55 $\pm$ 0.01 & 19.91 $\pm$ 0.47 & 11.38 $\pm$ 0.34 \\
Inr2Vec \cite{inr2vec,scale_equivariant_graph_metanetworks} & 23.69 $\pm$ 0.10 & 22.33 $\pm$ 0.41 & - \\
DWS \cite{dws_net} & 85.71 $\pm$ 0.57 & 67.06 $\pm$ 0.29 & 34.45 $\pm$ 0.42 \\
NFN$_{\text{NP}}$ \cite{perm_eq_neural_functionals,scale_equivariant_graph_metanetworks} & 78.50 $\pm$ 0.23 & 68.19 $\pm$ 0.28 & 33.41 $\pm$ 0.01 \\
NFN$_{\text{HNP}}$ \cite{perm_eq_neural_functionals,scale_equivariant_graph_metanetworks} & 79.11 $\pm$ 0.84 & 68.94 $\pm$ 0.64 & 28.64 $\pm$ 0.07 \\
NG-GNN \cite{kofinasgraph} & 91.40 $\pm$ 0.60 & 68.00 $\pm$ 0.20 & 36.04 $\pm$ 0.44 \\
ScaleGMN \cite{scale_equivariant_graph_metanetworks} & 96.57 $\pm$ 0.10 & 80.46 $\pm$ 0.32 & 36.43 $\pm$ 0.41 \\
ScaleGMN-B  \cite{scale_equivariant_graph_metanetworks} & 96.59 $\pm$ 0.24 & 80.78 $\pm$ 0.16 & 38.82 $\pm$ 0.10 \\
ProbeGen \cite{kahanadeep} & 98.40  $\pm$ 0.10 & 87.70 $\pm$ 0.30 & - \\
WT \cite{gielisse2025end}  &  91.38  $\pm$ 1.67 &  83.97  $\pm$ 1.38 &  43.78  $\pm$ 0.64 \\
MWT \cite{gielisse2025end} &  98.33  $\pm$ 0.11 &  89.41  $\pm$ 0.25 &  56.90  $\pm$ 0.29 \\
MWT-L \cite{gielisse2025end}  &  98.80  $\pm$ 0.06 & 90.43  $\pm$ 0.23   &   59.57  $\pm$ 0.52  \\
MWT-L* \cite{gielisse2025end}  &  98.91 $\pm$ 0.11 &  90.35 $\pm$ 0.20  &  58.90  $\pm$ 0.35  \\
\midrule
HMoE-WT & 97.87 $\pm$ 0.18 & 87.57 $\pm$ 0.19 & 55.44 $\pm$ 0.51  \\  

HMoE-MWT &{99.06} $\pm$ {0.15} & {90.72} $\pm$ {0.25} &  {65.01} $\pm$ {0.63}   \\  
HMoE-MWT-L  & 98.93 $\pm$ 0.21 &  90.30 $\pm$ 0.27 & 64.40 $\pm$ 0.50  \\  
\bottomrule
\end{tabular}

\label{tab:performance_toydatasets}
\end{table}

\begin{table}[t]
\centering
\caption{Imagenette results under standard model configurations.}
\small
\setlength{\tabcolsep}{4pt}
\renewcommand{\arraystretch}{1}
\begin{tabular}{lccccc}
\toprule
\textbf{Method} & \textbf{Accuracy (\%)} & \textbf{PSNR (dB)}  & \textbf{CLS} & \textbf{INR} & \textbf{FLOPs} \\
\midrule

WT \cite{gielisse2025end} 
  & 47.02 $\pm$ 0.37 & 22.47 $\pm$ 0.04 & 1.1M & 0.33M & 2.3G \\
\addlinespace[3pt]

MWT \cite{gielisse2025end} 
  & 56.78 $\pm$ 0.27 & 21.14 $\pm$ 0.02 & 1.1M & 0.33M & 2.3G \\
\addlinespace[3pt]

MWT-L \cite{gielisse2025end} 
  & 60.62 $\pm$ 0.37 & 22.31 $\pm$ 0.03 & 4.3M & 1.3M & 19G \\
\addlinespace[3pt]

\midrule

HMoE-WT 
  & 55.39 $\pm$ 0.70 & 22.49 $\pm$ 0.08 & 11M & 0.33M & 8.8G \\
\addlinespace[3pt]

HMoE-MWT 
  & 62.52 $\pm$ 0.60 & 21.47 $\pm$ 0.09 & 11M & 0.33M & 8.8G \\
\addlinespace[3pt]

HMoE-MWT-L 
  & {62.96 $\pm$ 0.55} & 23.01 $\pm$ 0.12 & 45M & 1.3M & 70G \\

\bottomrule
\end{tabular}

\label{tab:performance_imagenette_standard}
\end{table}

\begin{table}[!b]
\centering
\caption{Parameter-matched comparison on Imagenette. Rows are resized to match the classifier (CLS) parameter count (1.1M or 11M) while keeping the training protocol identical.}
\small
\setlength{\tabcolsep}{4pt}
\renewcommand{\arraystretch}{1}
\begin{tabular}{lccccc}
\toprule
\textbf{Method} & \textbf{Accuracy (\%)} & \textbf{PSNR (dB)}  & \textbf{CLS} & \textbf{INR} & \textbf{FLOPs} \\
\midrule

MWT \cite{gielisse2025end} 
  & 56.78 $\pm$ 0.27 & 21.14 $\pm$ 0.02 & 1.1M & 0.33M & 2.3G \\
\addlinespace[3pt]

HMoE-MWT-PM@1.1M 
  & {61.47 $\pm$ 0.32} & 21.38 $\pm$ 0.07 & 1.1M & 0.33M & 1.9G \\
\addlinespace[6pt]

HMoE-MWT 
  & {62.52 $\pm$ 0.60} & 21.47 $\pm$ 0.09 & 11M & 0.33M & 8.8G \\
\addlinespace[3pt]

MWT-PM@11M
  & 59.92 $\pm$ 0.35 & 21.45 $\pm$ 0.06 & 11M & 0.33M & 15.6G \\

\bottomrule
\end{tabular}
\label{tab:performance_imagenette_matched}
\end{table}

 \subsubsection{High-resolution datasets}

To demonstrate the scalability and effectiveness of our method, we evaluate it on two challenging datasets: Imagenette \cite{imagenette} and ImageNet-1K \cite{imagenet}. Imagenette is a 10-class subset of ImageNet that enables controlled large-scale experimentation. For fair comparison with prior work, we follow the same training protocol as \cite{gielisse2025end}, including identical spatial augmentations, hyperparameters, and four inner-loop optimization steps. To efficiently train on high-resolution images, we employ low coordinate subsampling rate of 0.1 during training. For all high-resolution experiments, we use SIREN INR with $\omega$ set to 30.

Table \ref{tab:performance_imagenette_standard} summarizes the results for Imagenette under standard model configurations. Our models consistently outperform the previous baselines \cite{gielisse2025end}. Notably, the HMoE-WT variant matches or exceeds the performance of the MWT, underscoring the intrinsic strength of our  MoE architecture. Furthermore, HMoE-MWT and HMoE-MWT-L improve classification accuracy beyond MWT-L, with HMoE-MWT-L achieving the highest classification accuracy.

Although our MoE design introduces additional parameters through the expert modules, the sparse routing mechanism ensures that only a single expert is active for any given token. As a result, the computational cost per token remains comparable to prior dense methods, while the overall model capacity increases due to the presence of multiple experts. In addition, to further illustrate the usefulness of our method, we compare MWT and HMoE-MWT under a matched parameter budget. We scale down HMoE-MWT by reducing the number of Transformer blocks from 10 to 8 and decreasing the FFN ratio from 1 to 0.25, yielding an HMoE-MWT model with 1.1M parameters, matching the original MWT. Conversely, we scale up the dense MWT by increasing the number of layers to 30 and setting the FFN expansion ratio to 8, resulting in approximately 11M parameters on Imagenette. While direct comparison of models with different architectures is inherently challenging, the results in Table \ref{tab:performance_imagenette_matched} demonstrate that our method maintains superior performance even in a strict parameter-matched comparison with dense baselines.

Next, we evaluate our largest model, HMoE-MWT-L, on the ImageNet-1K dataset using a coordinate subsampling rate of 0.01. SIREN width is set to 256. This model achieves a weight-space \textit{state-of-the-art} accuracy of 26.73\% while reconstructing images with a PSNR of 22.06 dB. A comparison with the results reported in \cite{gielisse2025end} is provided in Table~\ref{tab:imagenet}.

\begin{table}[t]
\centering
\caption{Our HMoE model outperforms the previous baseline on ImageNet-1K validation set.}
\small
\setlength{\tabcolsep}{4pt}
\begin{tabular}{p{2.4cm} ccc}
\toprule

\textbf{Method} & \textbf{Accuracy (\%)} & \textbf{PSNR (dB)} \\
\midrule
MWT-L \cite{gielisse2025end} &  24.11 &  21.78  \\
HMoE-MWT-L & {26.73}  & { 22.06 } \\

\bottomrule
\end{tabular}

\label{tab:imagenet}
\end{table}

\begin{figure}[!b]
      \centering
    \includegraphics[width=1\linewidth]{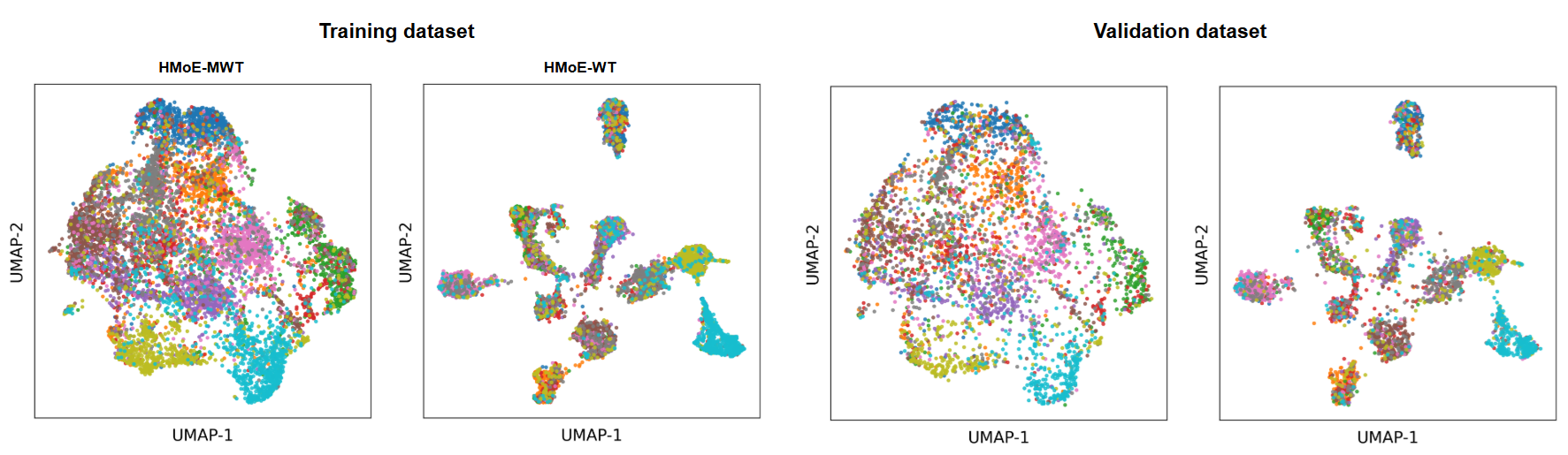}
    \caption{UMAP visualization of attribution-graph embeddings for HMoE-MWT (left) and HMoE-WT (right). Each point corresponds to an Imagenette sample and is colored by its ground-truth class. The meta-learned model produces smooth, continuous manifolds that remain consistent across training and validation splits, suggesting that meta-learning encourages distributed, reusable subcircuits shared across classes.}
    \label{fig:global_cluster}
\end{figure}

\subsection{Explainability in weight-space}

We focus on assessing the proposed method and explaining its performance on the Imagenette dataset.

\subsubsection{Weight manifold}

We compute Grad-CAG importance scores for each Imagenette sample in the training and validation splits, where each score reflects the contribution of an individual INR weight to the classifier’s output. After flattening the weighted parameters, we embed them into a two-dimensional space using UMAP \cite{mcinnes2018umap}. Figure~\ref{fig:global_cluster} shows the resulting manifolds for HMoE-MWT and HMoE-WT. Both models exhibit class-dependent structure in INR weight space, and applying the UMAP transformation learned on the training set to the validation set yields nearly identical patterns, indicating strong manifold-level generalization. A clear difference emerges between the meta-learned and non-meta-learned variants. HMoE-MWT forms smooth, continuous manifolds with substantial overlap across classes, suggesting that meta-learning promotes distributed, reusable subcircuits rather than rigid class-specific domains. In contrast, HMoE-WT produces fragmented, isolated clusters that align closely with class labels and shift between training and validation sets, consistent with its lower generalization performance.

\subsubsection{Weight pruning}

\begin{figure*}[t]
      \centering
    \includegraphics[width=0.9\linewidth]{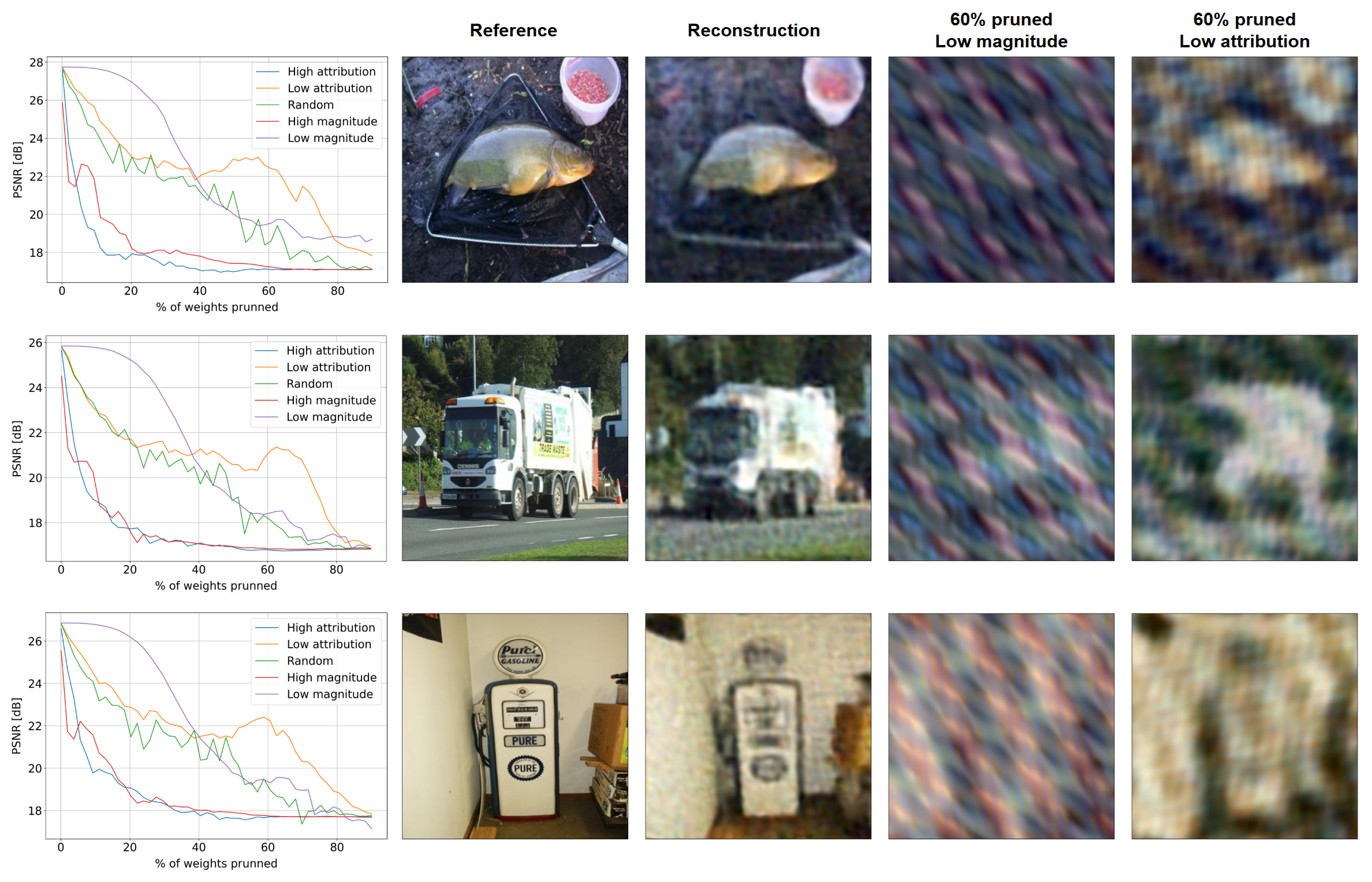}
    \caption{Functional pruning analysis of INR weight importance. PSNR curves show reconstruction degradation under different pruning strategies. Removing high-attribution or large-magnitude weights quickly destroys the reconstruction, while pruning low-attribution weights preserves quality longer.
Example reconstructions after pruning 60\% of the least important weights illustrate that attribution-based pruning retains coarse object structure better than magnitude pruning, suggesting the presence of functionally meaningful subcircuits.}
    \label{fig:prunning}
\end{figure*}

\begin{figure}
      \centering
    \includegraphics[width=0.9\linewidth]{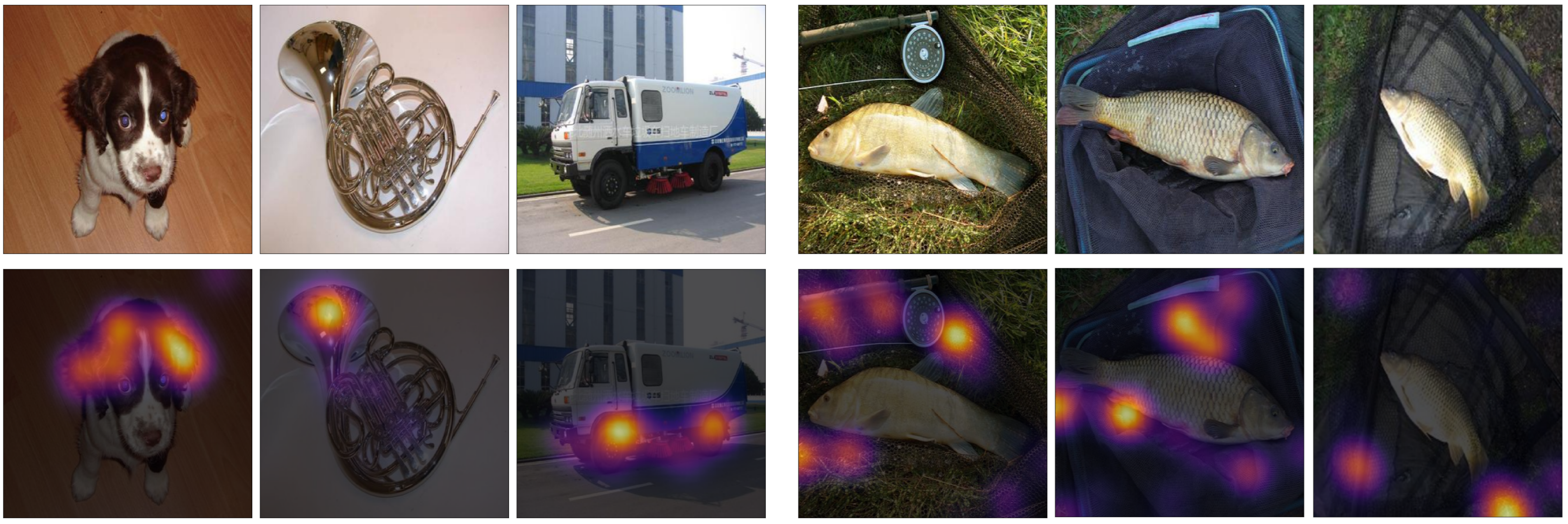}
    \caption{Attribution maps obtained via weight pruning. The first three examples show accurate and semantically meaningful attributions. The remaining three examples illustrate a typical failure mode of the dataset: the tench class is spuriously associated with the presence of the fishing net, a frequent co-occurring object.}
    \label{fig:attribution}
\end{figure}

We examine weight pruning guided by Grad-CAG importance scores. After pruning a chosen percentage of weights, we measure reconstruction quality via PSNR. We compare four strategies: pruning from least to most important (attribution–low), from most to least important (attribution–high), random pruning, and magnitude-based pruning that removes weights in order of their absolute value.

Figure \ref{fig:prunning} summarizes the results for HMoE-MWT on Imagenette. Pruning small-magnitude weights has little effect at first, whereas removing large-magnitude weights rapidly destroys the reconstruction. Attribution based pruning shows complementary behavior: removing the most important weights severely degrades reconstruction, confirming that they encode functionally critical structure. Pruning the least important weights initially resembles random pruning, but beyond roughly 40\% pruning, the PSNR remains higher than with magnitude based pruning. This suggests that class-relevant information is stored in subcircuits with diverse weight magnitudes, and attribution provides a more meaningful pruning criterion. Example reconstructions after pruning 60\% of the least important weights show that attribution-based pruning preserves coarse object structure, indicating that the identified weights correspond to functionally relevant subcircuits.

\subsubsection{Attribution maps}

Figure \ref{fig:prunning} shows that aggressive pruning driven by Grad-CAG retains coarse object identity, suggesting that pruning can be repurposed to generate spatial explanations. To obtain attribution maps analogous to Grad-CAM, we reconstruct the image after pruning 60\% of INR weights, compute the difference between the original and pruned reconstructions, invert it, downsample to 32×32, smooth with a Gaussian filter $(\sigma = 2)$, and upsample back to the original resolution \cite{fong2019understanding}. A nonlinear intensity remapping further highlights regions with the strongest pixel-level changes. The resulting maps (Fig.~\ref{fig:attribution}) indicate which spatial regions most influence the classifier. The first three examples illustrate accurate, semantically meaningful localization. The remaining cases reveal a characteristic dataset bias: the tench class is spuriously tied to the appearance of the fishing net, a frequently co-occurring object in Imagenette. This suggests that our method not only exposes informative regions but also surfaces dataset-level failure modes. 

To assess attribution quality, we manually annotated 200 segmentation masks (20 per class) from the Imagenette validation set and evaluated the maps using the pointing-game metric, which checks whether the maximal attribution point falls within the ground-truth mask. Our approach achieved a score of 0.35. In addition, we evaluated approaches based on magnitude based pruning and random pruning, which achieved scores of 0.22 and 0.15, respectively. These results demonstrate reasonable localization accuracy for our purely weight-space attribution technique.

\subsection{Ablations}

We perform an ablation study on CIFAR-10 to assess the components of our hierarchical architecture (Table~\ref{tab:ablations}). We focus on HMoE-MWT and compare several MoE configurations. Replacing the layer-wise MoE with a standard token-wise MoE (first three rows) leads to a noticeable drop in accuracy, showing that routing tokens by INR layer is important for capturing layer-level structure in the weight space. We also find that increasing the number of active experts does not improve performance, indicating that sparse routing is sufficient. We then vary the number of experts in both MoE stages. While additional experts slightly improve reconstruction PSNR, they do not yield higher classification accuracy. Overall, the results suggest that sparse, layer-aware MoE is well suited for INR weight-space classification, and that increasing expert capacity alone does not provide meaningful accuracy gains. Additional ablations, exploring the impact of selected hyper-parameters, are provided in the supplementary materials.

\begin{table}[t]
\centering
\caption{Ablation study evaluating the proposed hierarchical MoE architecture under different expert configurations. 
The first three rows correspond to replacing the layer-wise MoE with a standard token-wise MoE. 
These settings lead to lower accuracy, indicating that layer-wise routing provides a clear benefit for weight-space classification. }
\resizebox{0.9\columnwidth}{!}{
\begin{tabular}{ccccccc}
\toprule

TopK (Layer)  & \# Layer experts & TopK (Token)    &   \# Token experts   &   Acc [\%]  &   PSNR [dB]   &  Params  \\
\midrule
 1 & 4 & - & - & 63.21 & 33.86 & 11M \\
 - & - & 1 & 4 & 62.57 & 30.70 & 11M \\
 - & - & 2 & 4 & 62.54 & 31.14 & 11M \\
 - & - & 4 & 4 & 62.21 & 30.68 & 11M \\
 1 & 4 & 1 & 4 & \textbf{65.01} & 31.71 & 11M \\
 1 & 8 & 1 & 4 & 64.03 & 31.86 & 17M \\
 1 & 4 & 1 & 8 & 64.53 & \textbf{32.27} & 17M \\
 1 & 8 & 1 & 8 & 63.64 & 30.61 & 22M \\
\bottomrule
\end{tabular}
}

\label{tab:ablations}
\end{table}

%% file: sec/5_conclusion.tex
\section{Conclusions}

In this work, we introduced a Mixture-of-Experts architecture for learning in weight-space of implicit neural representations and demonstrated that our approach is well suited for processing the structured weight graphs produced by INRs. Our method achieves \textit{state-of-the-art} performance among weight-space models across multiple benchmarks and provides the first systematic explainability tools for this setting. Through attribution and pruning analyses, we identified specific INR subcomponents that contribute most to classification, offering new insight into how discriminative information is encoded in weight space.

Our approach also has limitations. Mixture-of-Experts models require longer training and higher computational cost than standard Transformers. Despite strong gains within the INR setting, performance still lags behind conventional pixel-based models on large-scale datasets such as ImageNet-1K.
Furthermore, the INR classification setting requires encoding data into this representation before downstream processing. We recognize that, while using meta-learning amortizes the computational cost of this representation, training one model per image remains a computational bottleneck. 
Additionally, our current experiments focus exclusively on image-based SIRENs to establish these weight-space design principles. While the HMoE framework naturally extends to diverse signals, testing its generalization across other INR architectures and domains (e.g., 3D NeRFs) is left for future work. Finally, our explainability analysis is a preliminary first step, focusing only on weight importance and pruning, leaving many aspects of weight-space interpretability unexplored.

Nonetheless, we believe these results mark a meaningful step toward more effective and interpretable weight-space learning, and we hope they inspire further work at the intersection of INRs, meta-learning, and model interpretability.

\section*{Acknowledgment}

This work was funded by the National Science Centre of Poland, grant number 2025/57/B/ST7/04045.

%% file: supp.tex
\title{Weight-Space Mixture-of-Experts for Implicit Neural Representation Classification} 

\titlerunning{Weight-space mixture-of-experts}
\author{Stanislaw Janik\inst{1}, Michal Byra\inst{1,2}}

\authorrunning{S. Janik and M. Byra}

\institute{Institute of Fundamental Technological Research, \\ Polish Academy of Sciences, Warsaw, Poland \and
Samsung AI Center, Warsaw, Poland\\
\email{\{sjanik,mbyra\}@ippt.pan.pl}}

\newcommand{\gradZero}{
g_{\theta, \alpha}^{\text{rec}} \leftarrow \nabla_{\theta, \alpha}\mathcal{L}_{\text{rec}}(f_\phi, \bf{x})
}
\newcommand{\gradOne}{
g_{\theta, \alpha}^{\text{cls}} \leftarrow  \nabla_{\theta, \alpha}\mathcal{L}_{\text{cls}}(\hat{y}, y)
}
\newcommand{\gradTwo}{
g_{\psi} \leftarrow \nabla_{\psi}(\mathcal{L}_{\text{cls}}(\hat{y}, y) * w_{\text{cls}} + \mathcal{L}_{\text{balance}} * w_{\text{balance}})
}

\newcommand{\gradThree}{
g_{\theta, \alpha} \leftarrow g_{\theta, \alpha}^{\text{rec}} + g_{\theta, \alpha}^{\text{cls}} * w_{\text{cls}} 
}

\maketitle

\section{Additional training details}
\label{sec:train_det}

To jointly train the SIREN implicit network and the weight Transformer, we utilize the meta-learning framework. Algorithm 1 illustrates this approach, where we additionally incorporate our hierarchical Mixture-of-Experts (HMoE) method. The meta-learning setup consists of two loops. The outer loop iterates over all training images, while the inner loop updates the shared learned initialization $\theta$. This initialization $\theta$ is optimized to serve as a shared starting point for efficiently fitting images from the target dataset. To fit an image and determine SIREN's parameter $\phi$, the inner loop is executed for $k$ steps. In addition, a separate learning-rate schedule $\alpha$ is learned for each inner-loop step, ensuring faster convergence. The inner loop is trained using the mean squared error, while the outer loop uses both the reconstruction and classification losses to update the initialization and the learning-rate schedule.

The classification loss, consisting of the cross-entropy loss and the balancing loss, is weighted by $w_{\text{task}}$, which is set to $0.01$ by default. Each HMoE block contains two routing mechanisms: a layer-wise MoE with $E_L$ experts and a token-wise MoE with $E_T$ experts. To encourage uniform expert utilization and prevent routing collapse, we adopt a standard load-balancing loss, which for the layer-wise MoE can be expressed as follows:
\[
\mathcal{L}^{\mathrm{layer}}_{\mathrm{balance}}
=
E_L \sum_{e=1}^{E_L} s_e\, P_e,
\]

\noindent where ${s}_j$ and ${P}_j$ are the assignment frequencies and 
mean gate probabilities for layer-level experts. The token-wise MoE computes an analogous loss:
\[
\mathcal{L}^{\mathrm{token}}_{\mathrm{balance}}
=
E_T \sum_{j=1}^{E_T} \tilde{s}_j\, \tilde{P}_j,
\]
where $\tilde{s}_j$ and $\tilde{P}_j$ are the assignment frequencies and 
mean gate probabilities for token-level experts. Each HMoE block returns the average of the two auxiliary losses,
\[
\mathcal{L}_{\mathrm{balance}}^{\mathrm{block}}
= 
\frac{1}{2}
\left(
\mathcal{L}^{\mathrm{layer}}_{\mathrm{balance}}
+
\mathcal{L}^{\mathrm{token}}_{\mathrm{balance}}
\right),
\]
and the total balancing loss $\mathcal{L}_{\mathrm{balance}}$is obtained by averaging this quantity across all MoE blocks and scaling it by a weighting factor $w_{\text{balance}}$, by default set to  $0.1$.

As mentioned in the Methods Section, our model operates on the difference $\phi - \theta$, where $\theta$ is the shared meta-learned initialization and $\phi$ denotes the image-specific INR weights obtained after the inner-loop optimization. Since this difference can contain elements with very small magnitude, we scale it by a factor $\lambda$. In practice, the classifier therefore operates on feature vectors of the form $\lambda(\phi - \theta)$, with $\lambda$ set to 500.

In our experiments, unless stated otherwise, for optimization we employ the AdamW optimizer. We use a batch size of 16 for models with SIREN width 128. The classifier is trained with a learning rate of 0.0001 and weight decay of 0.0001. For the inner-loop optimization, we use SGD without momentum, and we optimize the inner-loop learning rates using AdamW with a learning rate of 0.01.  For the SIREN network, we set $\omega = 10.0$ for MNIST, Fashion-MNIST, and CIFAR-10, and $\omega = 30$ for Imagenette and ImageNet-1K. Our Transformer classifier consists of 10 layers. 


\begin{algorithm}
	\caption{
            Task-Specific SIREN Meta-Learning.\\
            \small{
            For the inner-loop, we minimize a reconstruction loss $\mathcal{L}_{\text{rec}}$. We then optimize the initial SIREN parameters $\theta$ and learning rate schedule $\alpha$ such that $\phi$ does not only encode the image with high quality, but is also in a format that can be correctly classified by our classifier $c_{\psi}(\phi)$. Here, $f_{\phi}$ is our SIREN, $w_{\text{cls}}$ is a scalar that we can use to change the classifier influence on the meta-learning process, $w_{\text{balance}}$ is a scalar we can use to change the load balancing influence of MoEs. AdamW refers to the use of the AdamW  optimizer and similarly SGD refers to the SGD optimizer.
            }
    }\label{alg:metalearn}

	\begin{algorithmic}[1]
 
            \State Init random SIREN with parameters $\theta$
            \State Init learning rates $\alpha$ for all $k$ update steps $\alpha \in \mathbb{R}^{k \times |\theta|}$
            \While {not converged} \hfill \texttt{<outer loop>}
                \State Sample training image $\bf{x}$ with classification label $y$ 
                \State Set starting INR parameters to shared base $\phi = \theta$
                \For {$i=(1,2,\ldots, k)$} \hfill \texttt{<inner loop>}
                    \State $\phi \leftarrow \phi - \alpha_{i} \nabla_{\phi}\mathcal{L}_{\text{rec}}(f_\phi, \bf{x})$
                \EndFor
                \State Predict classification label and calculate the balancing loss $\hat{y}, \mathcal{L}_{\text{balance}} \leftarrow c_{\psi}(\phi)$
                \State Get $\theta, \alpha$ gradient $\gradOne$
                \State Get $\theta, \alpha$ gradient   $\gradZero$
                \State Combine $\theta, \alpha$ gradients $\gradThree$
                \State Update SIREN initialization $\theta \leftarrow \text{SGD}(\theta, g_{\theta})$
                \State Update learning rates $\alpha \leftarrow \text{AdamW}(\alpha, g_{\alpha})$ 
                \State Get $\psi$ gradient $\gradTwo$
                \State Update classifier $\psi \leftarrow \text{AdamW}(\psi, g_{\psi})$
            \EndWhile
	\end{algorithmic} 
\end{algorithm}

\section{Additional ablation studies}
\label{sec:add_abl}

\begin{table}[!t]
\centering
\caption{Ablation results and training times for the proposed HMoE-MWT model, trained on 80\% of the CIFAR-10 training set and evaluated on the remaining 20\%.} 
\resizebox{\columnwidth}{!}{
\begin{tabular}{p{3cm} ccccccc}
\toprule

\textbf{Ablation} & \textbf{Parameter} & \textbf{Accuracy} & \textbf{PSNR}   & \textbf{\#Params} & \textbf{\#Params} \\
    & & (\%) & (dB) & CLS & SIREN \\
\midrule

\multirow{4}{*}{\textbf{Task Influence} $w_{\text{task}}$}
 & 0.001 & 62.63 & 38.73 & 	 	11M & 	465K \\
 & \underline{0.01} &  63.14 & 30.39	  & 	11M	 	 & 465K	 \\
 & 0.1 & 58.54 & 22.98   & 11M & 	465K \\
 & 1 & 55.34 & 20.21 &  	11M & 	465K \\
 
\midrule

 \multirow{5}{*}{\textbf{Inner-Loop Steps $k$ }}
 & 1 & 57.84 & 	23.01  & 		 	11M & 	133K \\
& 2 & 60.23 & 	28.17  & 		 	11M & 	199K \\
& 4 & 62.22 & 	29.72  & 	 		11M & 	332K \\
& \underline{6} &63.14 & 30.39	  & 	 11M	 	 & 465K	 \\

\midrule

 \multirow{3}{*}{\textbf{Siren Depth }}
 & 2 & 62.08 & 	31.11  & 			11M & 	234K \\
& \underline{4} & 63.14 & 30.39	  & 	 11M	 	 & 465K	 \\
& 6 & 64.10 & 	31.19  & 	 		11M & 	697K \\

\midrule

 \multirow{3}{*}{\textbf{Siren $\omega$ }}
 & 5 & 63.16 & 	30.4  & 	 		11M & 	465K \\
& \underline{10} &  63.14 & 30.39	  & 	 11M	 	 & 465K	 \\
& 15 & 62.86 & 	30.43  & 	 		11M & 	465K \\



\midrule

 \multirow{3}{*}{\textbf{Transformer Depth }}
 & 5 & 62.04 & 	31.23  & 			6M & 	465K \\
& \underline{10} &  63.14 & 30.39	  & 	 11M	 	 & 465K	 \\
& 15 & 63.59 & 	31.13  	 & 	17M	 & 	465K \\

\midrule

 \multirow{3}{*}{\textbf{Balancing Weight}}
 & 0.01 & 56.40 & 30.91 & 			11M & 	465K \\
& \underline{0.1} & 63.14 & 30.39	  & 	 11M	 	 & 465K	 \\
& 1 & 63.51 & 	31.19  & 	 		11M & 	465K \\

\midrule
\multirow{4}{*}{\textbf{Subsampling Rate }}
 & \underline{1.0} & 63.14 & 30.39	  & 	 11M	 	 & 465K	 \\
 & 0.1 & 61.08 & 	24.27	 &  	11M & 	465K \\
 & 0.01 & 55.36 & 	18.97 &   	11M & 	465K \\
 & 0.001 & 11.59 & 	16.15 & 	 		11M & 	465K \\

\bottomrule
\end{tabular}
}

\label{tab:ablation_additional}
\end{table}

Ablation studies in the Experiments Section examined the impact of varying the number of experts on classification performance. Here, we further investigate the influence of selected training parameters. We evaluate the HMoE-MWT model with four experts in both the layer-wise and token-wise MoE blocks, using TopK with $k$=1 routing in both cases. The results are summarized in Table~\ref{tab:ablation_additional}. Underlines indicate the model trained with the default settings used in our experiments, ensuring a fair comparison with the previous baseline.

Similar to the behavior observed for MWT, increasing the classification loss weight $w_{\text{task}}$ does not improve accuracy. The highest accuracy is achieved with $w_{\text{task}} = 0.01$, while both lower and higher values lead to reduced performance. A plausible explanation is that maintaining high-quality reconstructions yields stronger gradients for meta-learning, whereas emphasizing classification too strongly may hinder the formation of useful INR representations.

As in the original MWT, using a higher subsampling rate leads to improved reconstruction quality and higher classification accuracy. The subsampling rate refers to the fraction of pixel coordinates used to train each INR during the inner-loop fitting step, and we adjust this rate depending on the resolution of the dataset. Lower subsampling rates reduce computational cost while still providing enough signal for effective meta-learning. Overall, the hyperparameter trends of our model closely follow those reported for MWT, indicating that HMoE-MWT retains similar optimization characteristics while improving classification performance.

\section{Expert Routing}
\label{sec:expert_routing}

To further validate the design of our hierarchical MoE architecture, we conducted an additional experiment analyzing expert utilization. Specifically, we calculated the average activation frequency of each expert across the entire Imagenette validation set. 

\begin{figure}[ht]
    \centering
    \includegraphics[width=0.9\linewidth]{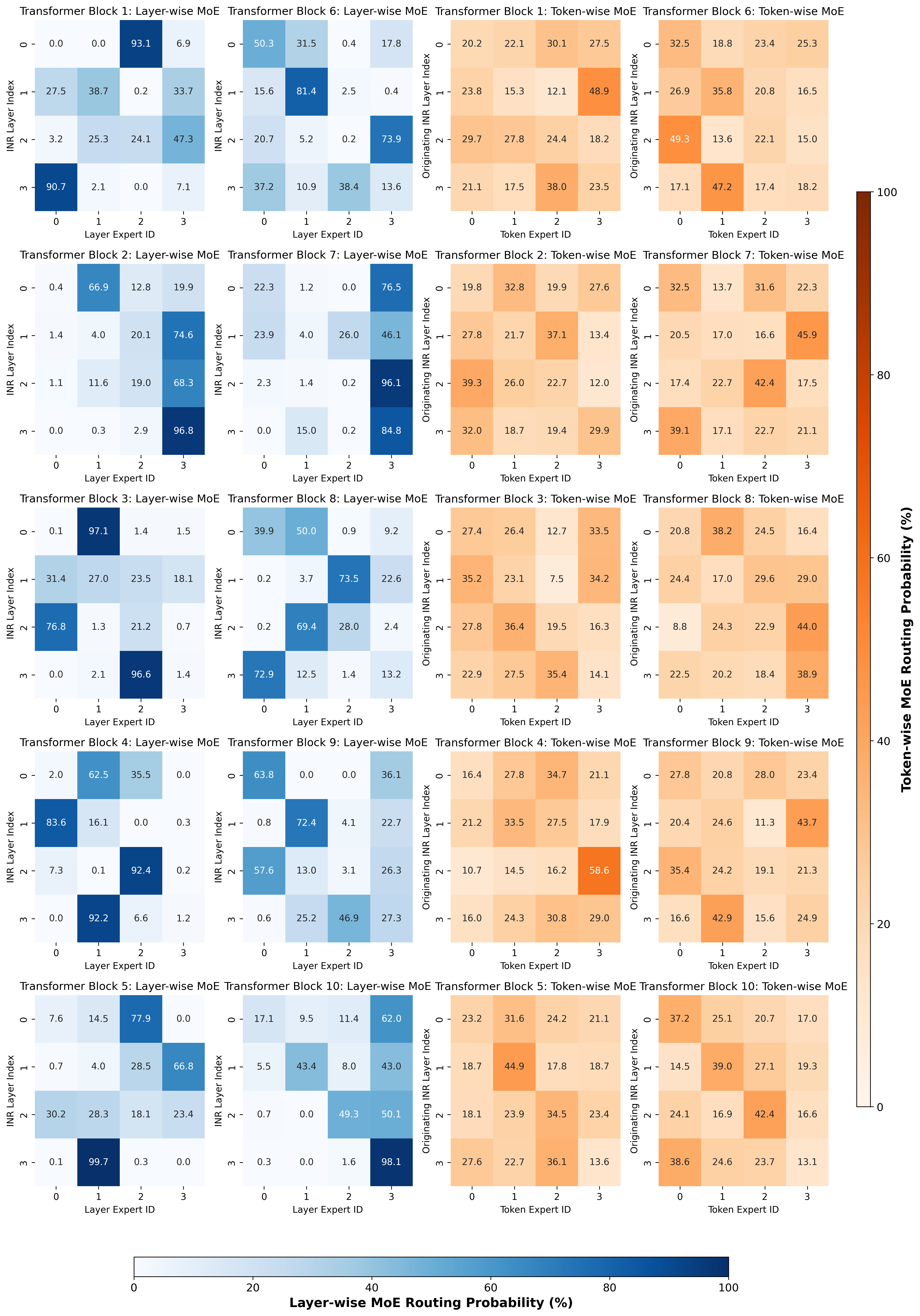}
    \caption{Expert routing analysis over the full Imagenette validation dataset. Layer-wise MoE (left) shows clear layer-expert specialization with dominant experts per INR layer validating that different INR layers encode functionally distinct information. Token-wise MoE (right) exhibits balanced, distributed routing (17--49\% range), enabling content-adaptive processing. The consistency across the validation set demonstrates stable generalization without routing collapse.}
    \label{fig:experts}
\end{figure}

As shown in Fig.~\ref{fig:experts}, the layer-wise MoE exhibits dominant experts for specific INR layers within each block, whereas the token-wise MoE displays a more evenly distributed expert utilization. This routing behavior suggests that coarse, layer-specific structural information is effectively captured by the layer-wise experts. Consequently, this allows the token-wise experts to focus on fine-grained, layer-independent features. Furthermore, the active utilization of all experts across the network confirms that our load-balancing loss successfully prevents routing collapse.

